\documentclass[letterpaper]{article} % DO NOT CHANGE THIS
\usepackage{aaai2026}  % DO NOT CHANGE THIS
\usepackage{times}  % DO NOT CHANGE THIS
\usepackage{helvet}  % DO NOT CHANGE THIS
\usepackage{courier}  % DO NOT CHANGE THIS
\usepackage[hyphens]{url}  % DO NOT CHANGE THIS
\usepackage{graphicx} % DO NOT CHANGE THIS
\usepackage{natbib}  % DO NOT CHANGE THIS AND DO NOT ADD ANY OPTIONS TO IT
\usepackage{caption} % DO NOT CHANGE THIS AND DO NOT ADD ANY OPTIONS TO IT
\usepackage{algorithm}
\usepackage{algorithmic}

\usepackage{newfloat}
\usepackage{listings}
\DeclareCaptionStyle{ruled}{labelfont=normalfont,labelsep=colon,strut=off} % DO NOT CHANGE THIS
\floatstyle{ruled}
\newfloat{listing}{tb}{lst}{}
\floatname{listing}{Listing}
\usepackage{amsmath} 
\usepackage{amssymb}
\usepackage{mathtools}
\usepackage{multirow}
\usepackage[table]{xcolor}
\usepackage{booktabs}
\definecolor{lightred}{rgb}{1, 0.55, 0.45}    % 浅红（红偏多，绿蓝稍低）
\definecolor{lightblue}{rgb}{0.45, 0.6, 1}   % 浅蓝（蓝偏多，绿稍高增加柔和度）
\definecolor{lightgreen}{rgb}{0, 0.733, 0.314}  % 浅绿（绿偏多，红蓝稍低）
\definecolor{ourpink}{rgb}{1,0.8,0.8}
\definecolor{ourbaseline}{rgb}{0.95,0.95,0.95}
\definecolor{ourgray}{rgb}{0.88,0.88,0.88}
\definecolor{darkgreen}{RGB}{0, 168, 64} % 定义一个暗绿色

\usepackage{marvosym}
\usepackage{multirow}   % 支持 \multirow 命令
\usepackage{booktabs}   % 支持 \toprule、\bottomrule 等

\title{\raisebox{-0.3\height}{\includegraphics[width=0.05\textwidth]{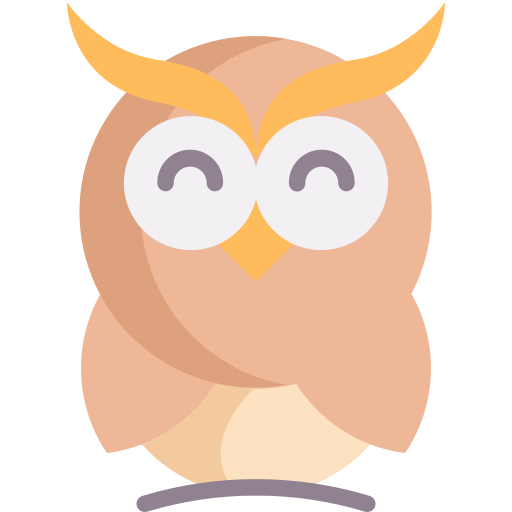}}\hspace{-0.2em} OwlCap: Harmonizing Motion-Detail for Video Captioning via \\HMD-270K and Caption Set Equivalence Reward}

\author{Chunlin Zhong \textsuperscript{\rm 1}\equalcontrib,
Qiuxia Hou \textsuperscript{\rm 2}\equalcontrib,
Zhangjun Zhou \textsuperscript{\rm 1}\equalcontrib, 
Yanhao Zhang\textsuperscript{\rm 2}\thanks{Corresponding Authors.},\\
Shuang Hao\textsuperscript{\rm 1,3},
Haonan Lu\textsuperscript{\rm 2},
He Tang\textsuperscript{\rm 1}\textsuperscript{\textdagger},
Xiang Bai\textsuperscript{\rm 1}\textsuperscript{\textdagger}
} 
\affiliations{\textsuperscript{\rm 1}School of Software Engineering, Huazhong University of Science and Technology,
Wuhan, China \\
\textsuperscript{\rm 2}OPPO AI Center, OPPO Inc., China\\
\textsuperscript{\rm 3}School of Life Science and Technology, Xi’an Jiaotong University, Xi’an, China\\
\{clzhong, hetang, xbai\}@hust.edu.cn,
\{houqiuxia, zhangyanhao\}@oppo.com
}
\usepackage{bibentry}
\makeatletter\def\@listi{\leftmargin\leftmargini \topsep .5em \parsep .5em \itemsep .5em}
\def\@listii{\leftmargin\leftmarginii \labelwidth\leftmarginii \advance\labelwidth-\labelsep \topsep .4em \parsep .4em \itemsep .4em}
\def\@listiii{\leftmargin\leftmarginiii \labelwidth\leftmarginiii \advance\labelwidth-\labelsep \topsep .4em \parsep .4em \itemsep .4em}\makeatother

\newcounter{checksubsection}
\newcounter{checkitem}[checksubsection]

\begin{document}

\maketitle

\begin{abstract}
Video captioning aims to generate comprehensive and coherent descriptions of the video content, contributing to the advancement of both video understanding and generation.
However, existing methods often suffer from motion-detail imbalance, as models tend to overemphasize one aspect while neglecting the other. This imbalance results in incomplete captions, which in turn leads to a lack of consistency in video understanding and generation.
To address this issue, we propose solutions from two aspects: 
1) Data aspect: we constructed the Harmonizing Motion-Detail 270K (\textbf{HMD-270K}) dataset through a two-stage pipeline: Motion-Detail Fusion (MDF) and Fine-Grained Examination (FGE). 
2) Optimization aspect: We introduce the Caption Set Equivalence Reward (\textbf{CSER}) based on Group Relative Policy Optimization (GRPO). 
CSER enhances completeness and accuracy in capturing both motion and details through unit-to-set matching and bidirectional validation.
Based on the HMD-270K supervised fine-tuning and GRPO post-training with CSER, we developed \textbf{OwlCap}, a powerful video captioning multi-modal large language model (MLLM) with motion-detail balance.
Experimental results demonstrate that OwlCap achieves significant improvements compared to baseline models on two benchmarks: the detail-focused VDC (+4.2 Acc) and the motion-focused DREAM-1K (+4.6 F1).
The HMD-270K dataset and OwlCap model will be publicly released to facilitate video captioning research community advancements.
\end{abstract}

% \begin{links}
%     \link{Code and Datasets}{https://aaai.org/example/code}
% \end{links}

\section{Introduction}
Video captioning is intended to generate comprehensive and detailed descriptions based on video content. 
Compared with image captioning, video captioning must depict static details while simultaneously capturing temporal motion specific to video.
It has become an important research domain in multimodal large language models (MLLMs) whose advances significantly impact other fields, including streaming video understanding~\cite{xiong2025streaming}, text-to-video generation~\cite{bar2024lumiere}, and video editing~\cite{chai2023stablevideo}.
\begin{figure}[t]
\centering
\includegraphics[width=0.96\linewidth]{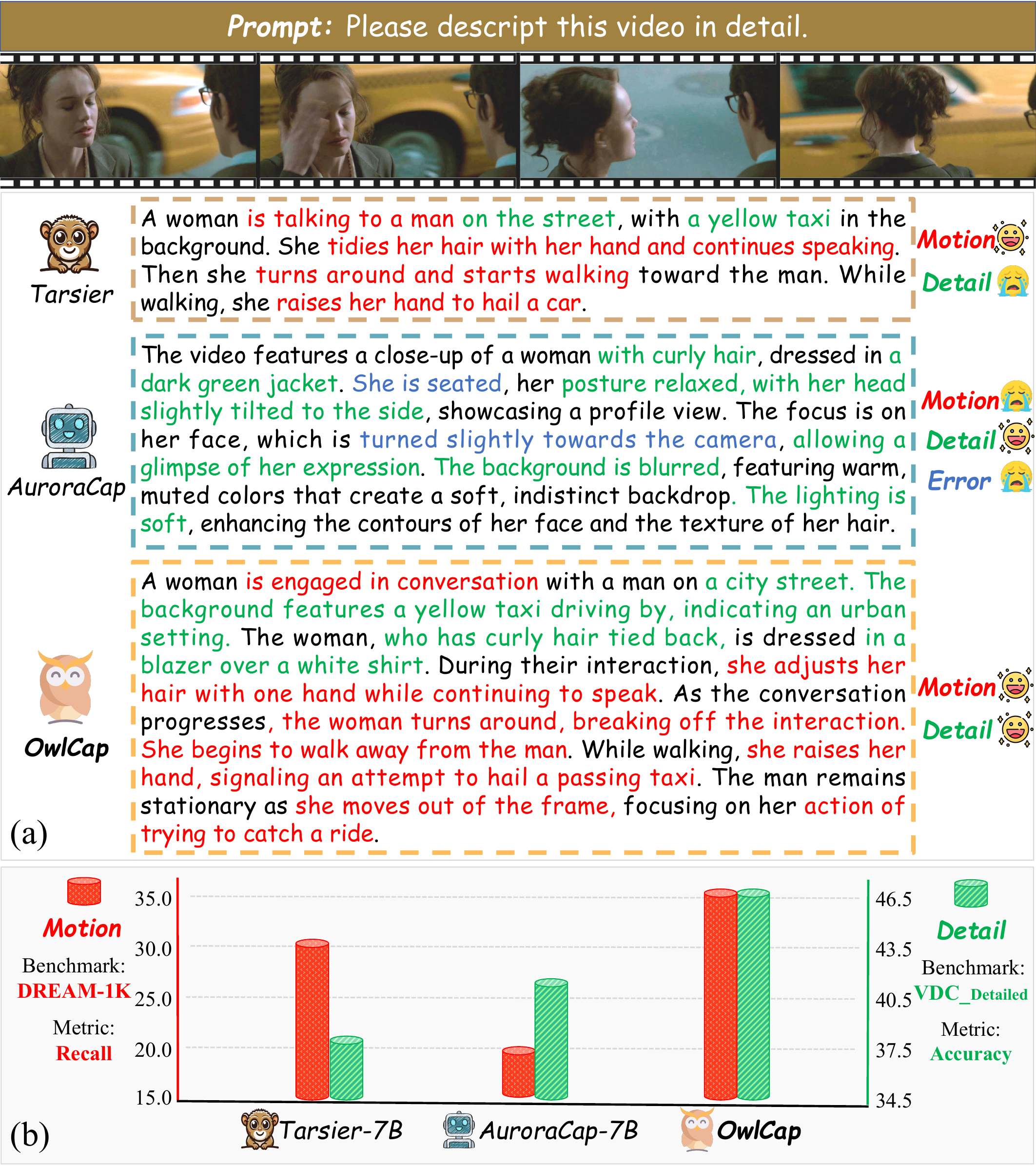}    
\caption{Video captioning encourages capturing both motion and detail. Qualitative (a) and quantitative (b) comparisons show that OwlCap manages to cover both aspects. 
% In (a) AuroraCap's errors are highlighted in blue.
}
\label{fig:motivation}
\end{figure}
With the development of MLLMs, some studies~\cite{chai2025auroracap,yuan2025tarsier2} have focused on improving performance in video captioning across two key dimensions: detail and motion.
% With the development of MLLMs, some studies have focused on improving model performance in Video  Captioning.  
AuroraCap~\cite{chai2025auroracap} enhances detail extraction capabilities by retraining on task-specific datasets, generating more detailed captions compared to traditional video captioning models. Tarsier~\cite{wang2024tarsier} strengthens motion and event capture in video modalities via large-scale pre-training and 150K human-annotated action-oriented video captions. 
However, while enhancing one aspect, they often overlook the other, 
optimizing for either detail characterization or motion capture in isolation, rather than achieving joint optimization of both dimensions:
% optimizing for detail characterization or motion capture in isolation and failing to achieve joint optimization of both dimensions: 
(1) as shown in Figure~\ref{fig:motivation}(a), Tarsier typically focuses on outputting motion information in videos, while AuroraCap emphasizes detail presentation, both neglecting the other aspect; (2) as illustrated in Figure~\ref{fig:motivation}(b), existing models tend to excel in specific benchmarks: Tarsier performs well on the motion-focused DREAM-1K benchmark~\cite{wang2024tarsier}, while AuroraCap excels on the detail-focused Video Detailed Captions (VDC) benchmark~\cite{chai2025auroracap}. 
In summary, existing MLLMs fail to balance motion and detail in video captioning. This imbalance results in incomplete captions, which in turn leads to a lack of consistency in video understanding and generation.

\begin{figure}[t]
\centering
\includegraphics[width=0.97\linewidth]{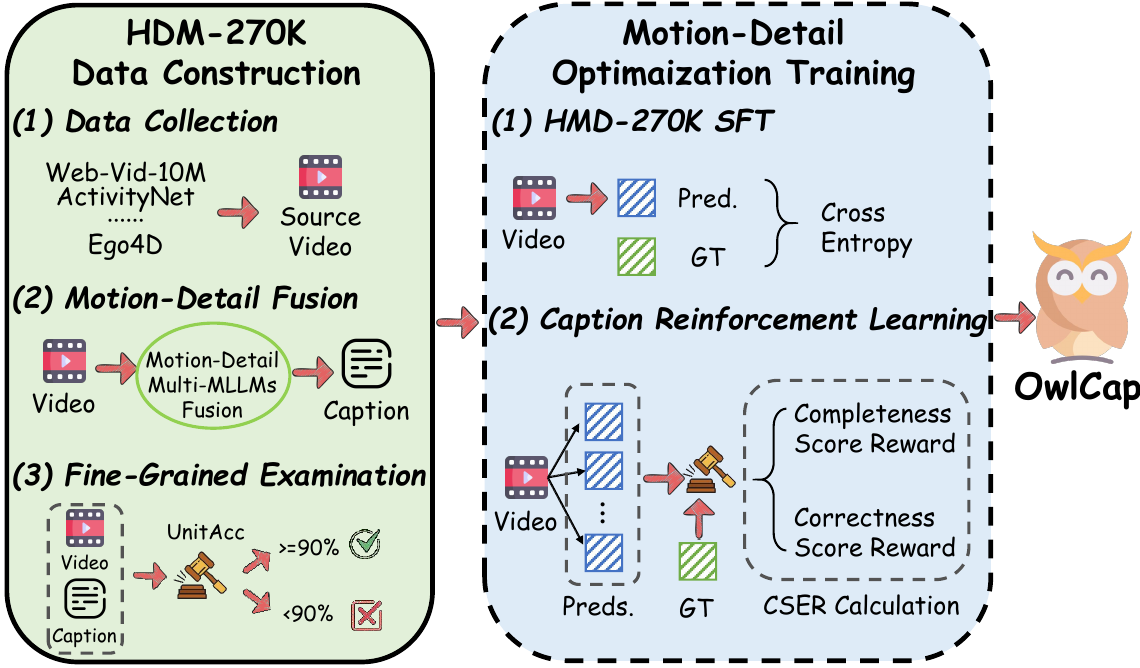}    
\caption{Pipeline for HMD-270K and OwlCap creation. 
}
\label{fig:overall}
\end{figure}
To address this, as shown in Figure~\ref{fig:overall}, we optimize motion-detail from two aspects: 

\textit{1) Data aspect}: 
We collect videos from various open-source datasets and construct the Harmonizing Motion-Detail 270K (\textbf{HMD-270K}) dataset using a two-stage pipeline. 
First, Motion-Detail Fusion (MDF) 
leverages distinct MLLMs to separately extract motion and detail information, then integrates them into a fused caption.
Second, Fine-Grained Examination (FGE) validates captions by decomposing them into units for individual verification.
As shown in Table 1, we calculated words and verbs per second of video in each dataset, and used these statistics to reflect the granularity of detail and motion in the captions. Based on these two statistics, we designed a Motion-Detail Balance (MDB) to measure the balance of the datasets.
These data indicate that HMD-270K contains more balanced and comprehensive motion-detail descriptions.

\textit{2) Optimization aspect}: 
We first perform Supervised Fine-Tuning (SFT) using the HMD-270K dataset, enabling the model to acquire motion-detail balance capabilities through training on this dataset. Furthermore, we introduce the Caption Set Equivalence Reward (\textbf{CSER}) based on the Group Relative Policy Optimization (GRPO)~\cite{shao2024deepseekmath}. Guided by set equivalence theory~\cite{cantor1874ueber}, CSER encourages the model to pursue completeness and correctness of captions through a unit-to-set matching and bidirectional validation strategy, thereby harmonizing the capture of motion and detail information.
Compared with VideoCap-R1’s Event Score~\cite{meng2025videocap}, which evaluates the single-sided ground-truth event entailed by the predicted caption, CESR adopts a more fine-grained and bidirectional approach to optimize both correctness and completeness of captions.

Based on HMD-270K SFT and  
CSER, we develop \textbf{OwlCap}, a powerful video captioning MLLM that captures both motion and detail information in videos. 
As shown in Figure 1(b), OwlCap outperforms in both the motion-focused DREAM-1K and the detail-focused VDC benchmarks simultaneously, achieving an accuracy of 46.8\% on VDC\_Detailed and a Recall of 35.3\% on DREAM-1K.
Furthermore, OwlCap outperforms existing video captioning methods in the Text-to-Video (T2V) generation task.
Our key contributions are summarized as follows:
\begin{itemize}
    \item We introduce HMD-270K, a large-scale dataset comprising 270K video-caption pairs that contain comprehensive motion-detail information, constructed via a designed two-stage pipeline.
    \item To address the lack of fine-grained modeling in video captioning methods, we introduce CSER, a reward function that employs unit-to-set matching and bidirectional validation to optimize completeness and correctness of generated captions.
    \item We propose OwlCap, which uses HMD-270K SFT and reinforcement learning with CSER, to enable balance motion-detail caption generation, and our approach outperforms all existing models on mainstream video captioning benchmarks.
\end{itemize}
\begin{figure*}[t]
  \centering
  \includegraphics[width=1\linewidth, keepaspectratio]{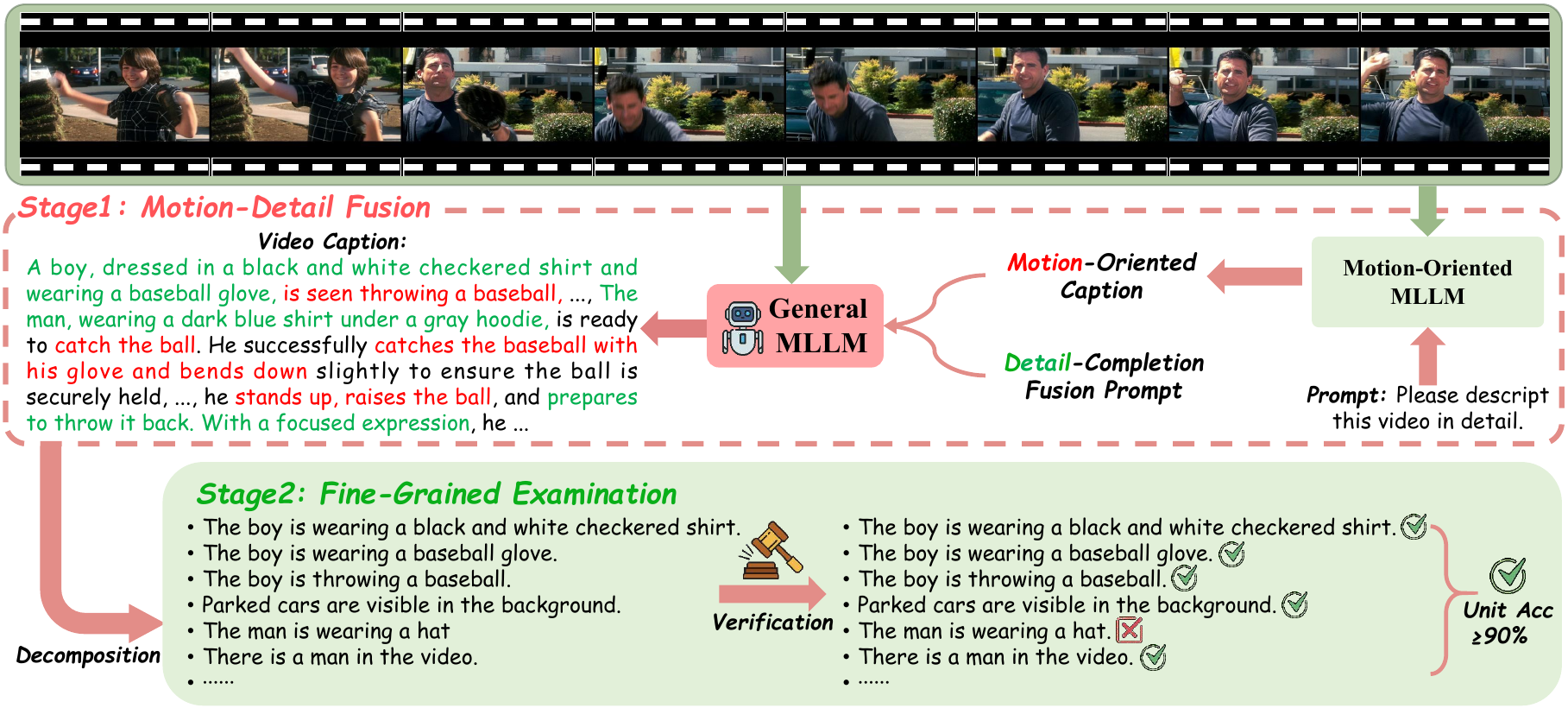}
  \caption{Pipeline for generating high-quality video captions that simultaneously accommodate motion and detail information.}
  \label{fig2:dataset}
\end{figure*}
\section{Related Work}
\subsection{Video Captioning}
Traditional video captions generate a brief text to summarize the main content of a video~\cite{yan2022videococa,yang2023vid2seq}. 
With the advancement of MLLMs, achieving more comprehensive and coherent video captioning has emerged as a key direction for model optimization~\cite{chen2024sharegpt4video, yang2024vript,chai2025auroracap}.
For instance, ShareGPT4Video~\cite{chen2024sharegpt4video} leverages detailed video captions generated by GPT-4V and trains the model using a differential video captioning strategy, enabling it to better describe detailed information. Taesier~\cite{wang2024tarsier} undergoes pre-training on 40 million video-text pairs and further fine-tunes with 150K manually annotated event-focused data, thereby enhancing the model's motion ability. However, these studies fail to effectively balance motion and detail. 
In contrast, our work addresses this issue by generating large-scale motion-detail balance video-caption pairs through a two-stage pipeline and designing a dedicated caption reward for video captioning.
% In contrast, our work addresses this issue by generating large-scale motion-and-appearance aware caption pairs through the HAE Pipeline and designing a dedicated reward function for Video Detailed Captioning.
\subsection{Reinforcement learning for MLLMs}
Recently, numerous studies have begun exploring the application of reinforcement learning to multimodal tasks~\cite{liu2025visual,feng2025video,li2025videochat}: Visual-RFT~\cite{liu2025visual} employs a GRPO-based strategy for various visual perception tasks, improving model performance under limited data. Similar efforts have emerged in the video understanding domain. Video-R1~\cite{feng2025video} utilizes temporal-based T-GRPO in multiple video tasks, enhancing the model's video understanding capability by integrating GRPO with temporal information. VideoCap-R1~\cite{meng2025videocap} attempts to use event coverage as a reward function to improve video caption quality, but its calculation method is overly simplistic, failing to adequately verify the correctness and completeness of captions. 
Our work further optimizes the model's captioning ability based on GRPO through more fine-grained unit division and more accurate calculation of rewards for correctness and completeness.
% \section{Harmonizing Motion-Detail Dataset}
\section{HMD-270K Dataset}
% Constructing large-scale video caption datasets enhances the captioning capabilities of MLLMs. 
To meet practical needs, we introduce a new dataset for video captioning, Harmonizing Motion-Detail (HMD-270K), which contains 270K video-caption pairs capturing both video motions and detailed descriptions. We will first introduce the dataset collection, then illustrate the construction pipeline of HMD-270K, and finally present its statistics.
\subsection{Dataset Collection}
To ensure the diversity of the dataset, we take both dynamic and static information of videos into account during the video collection process. 
Following Tarsier2-Recap-585K~\cite{yuan2025tarsier2}, we collected a large pool of videos drawn from the open-source datasets such as WebVid-10M~\cite{bain2021frozen}, ActivityNet~\cite{krishna2017dense}, and Ego4D~\cite{grauman2022ego4d}. A more detailed breakdown of video sources is provided in the Appendix.
These datasets cover multiple domains, including wildlife, movies, cooking, sports, news, and TV programs, providing a solid foundation for understanding various real-world scenarios. 
\textbf{Note that the videos used in the evaluation benchmark are not included in the HMD-270K.} 
\subsection{Dataset Construction Pipeline}
As shown in Figure \ref{fig2:dataset}, we designed an information fusion and filtering pipeline that leverages open-source video captioning MLLMs to generate motion-detail balanced captions through Motion-Detail Fusion (MDF) and Fine-Grained Examination (FGE) stages.
The MDF stage completes the captions, while the FGE stage verifies their accuracy.\\
% The MDF stage fuses captions, while the FGE stage verifies their accuracy.\\
\textbf{Motion-Detail Fusion Stage.} 
The defining characteristic of video captioning lies in two aspects: compared with image captioning, videos contain comprehensive temporal-related motion information specific to video; Furthermore, video captioning requires detailed descriptions of static elements—including object attributes, environmental features, and scene ambiance. To optimize these aspects during dataset construction, as shown in Figure \ref{fig2:dataset}, we first input the video into the motion-oriented MLLM Tarsier~\cite{wang2024tarsier}. Guided by a prompt designed to focus on temporal motion description, we generate a caption centered on temporal motion. After generation, this motion-focused caption and the original video are jointly fed into the general multimodal large model Qwen2.5-VL-72B~\cite{bai2025qwen2}. Through a prompt that guides it to supplement static details, we ultimately generate a video caption that simultaneously contains temporal motion and static details. The full prompt is provided in the Appendix.\\
\textbf{Fine-Grained Examination Stage.}
Compared to other video understanding tasks, video captioning requires describing all video elements.
This makes it highly challenging to directly use MLLMs to determine whether a generated caption aligns with the video content. 
To address this, we design a dedicated ``examination" to evaluate the consistency between the generated caption and the video content. 
As illustrated in Figure~\ref{fig2:dataset}, we decompose the video caption into indivisible units and validate each unit against the video using the Judge Model (InternVL2.5-78B in our pipeline). 
The motivation behind FGE stems from the fact that this decomposition reduces ambiguity and markedly enhances the transparency and interpretability~\cite{ye2025painting}. 
% As illustrated in Figure~\ref{fig2:dataset}, our approach involves decomposing the video caption into multiple indivisible units. These units are then validated one-by-one against the video using a Judge Model (InternVL2.5-78B in our pipeline). 
Finally, We set a 90\% unit accuracy (Unit Acc) threshold for filtering video-caption pairs to address two challenging scenarios: subjective descriptions in captions (e.g., "exuding a warm atmosphere") that lead to inconsistent judgments, and ambiguous descriptive targets in videos (e.g., "speaking" vs. "arguing") that cannot be accurately distinguished using only video frames. This threshold allows for some model fault tolerance.
To address potential bias, Appendix includes three complementary checks: (i) a cross-model consistency check that compares unit-level labels from our Judge model (InternVL2.5-78B) with those from GPT-4o on the same 500 FGE samples; (ii) a sensitivity analysis that evaluates model performance when use different FGE threshold; and (iii) a human re-evaluation of 100 automatically rejected samples to confirm the validity of the 90\% cutoff.
\begin{table}[!t]  % 
    \centering
    \renewcommand{\arraystretch}{0.95}
    \renewcommand{\tabcolsep}{1.9mm}
    \begin{tabular}{l|ccc}  % 调整列结构，增加三列
    \toprule
    \multirow{2}{*}{{Dataset}}  
    & Words&Verbs &\multirow{2}{*}{{MDB~$\uparrow$ }} \\  
    & Per second&Per second  &\\
    \hline
    Panda-70M 
     &~~1.5
      &0.2
        &0.22\\
    ShareGPT4Video &10.7
    &1.3  &0.54\\
    Vript &13.0
    &1.4 &0.50 \\
    Tarsier-585K &~~5.2
    &0.8 &0.49 \\
    \cellcolor{ourgray}HMD-270K &
    \cellcolor{ourgray}\textbf{13.9}&
     \cellcolor{ourgray}\textbf{2.0}&
      \cellcolor{ourgray}\textbf{0.68}\\
    \bottomrule
    \end{tabular}
    \caption{Comparison of different video caption datasets. MDB stands for Motion-Detail Balance, its formula and detailed introduction can be found in ``Data Statistics''.}
    \label{tab:dataset}
\end{table}
\begin{figure}[!t]
  \centering
  \includegraphics[width=0.81\linewidth, keepaspectratio]{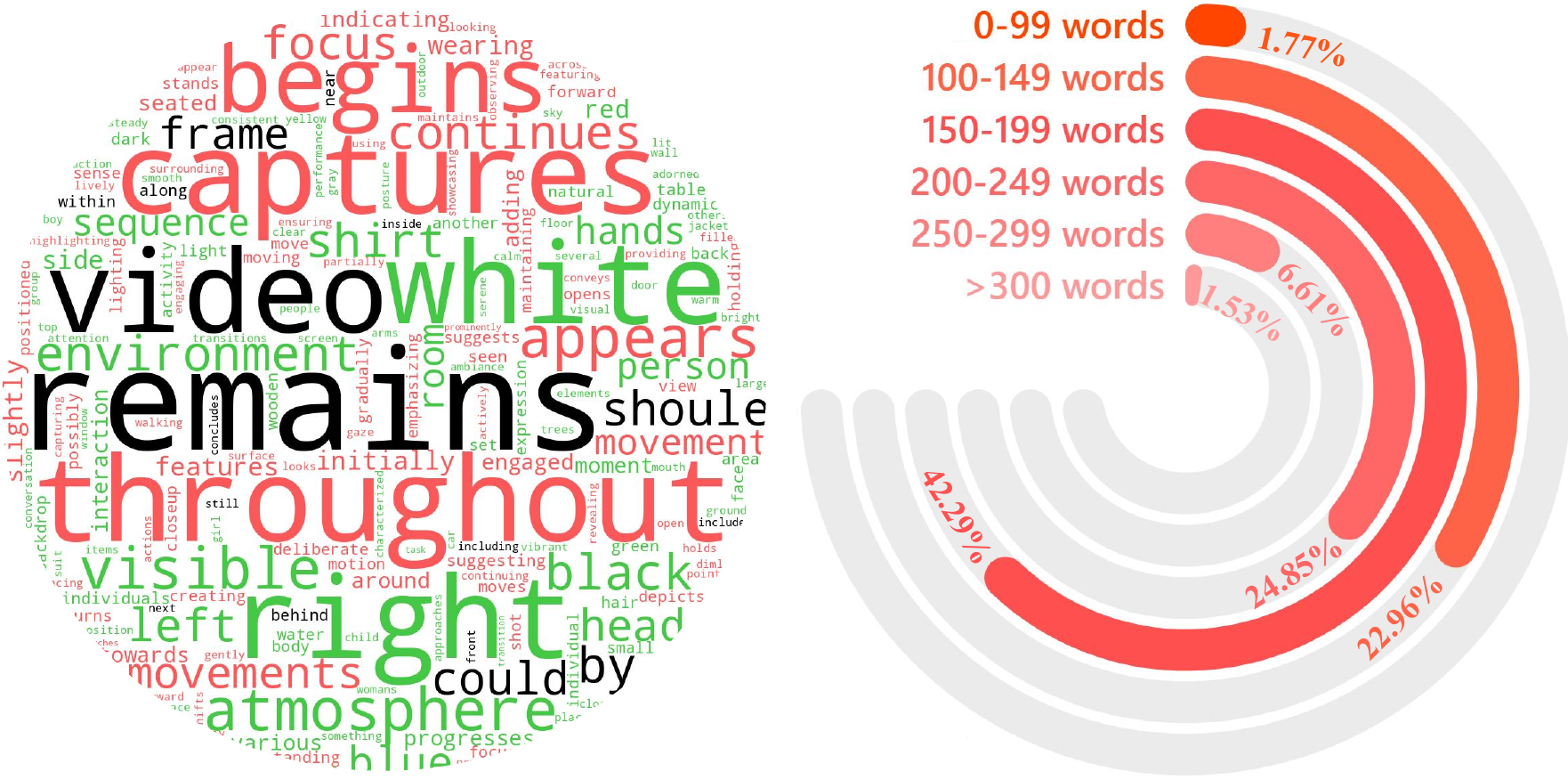}
    \caption{Word cloud (left) and length distribution (right) of captions in HMD-270K.}
  \label{fig3:wordStatistic}
\end{figure}
\subsection{Data Statistics}
The HMD-270K dataset, constructed via a two-stage pipeline, captures both motion and detailed information in videos. Table~\ref{tab:dataset} provides statistics on the number of words and verbs per second for various video captioning datasets, indicating the comprehensiveness of their descriptions of motion and detail. To quantify the balance between motion and detail, we introduced the Motion-Detail Balance (MDB) metric, calculated using the following formula:
\begin{equation}
    \text{MDB} = \left(1 - \frac{w - v}{w + v}\right) \cdot \log(w + 1),
\end{equation}
where $w$ and $v$ denote the number of caption words and verbs per second, respectively.
MDB comprehensively accounts for both Motion and Detail in captions. Specifically, 
($1 - \frac{w - v}{w + v}$) 
measures the proportion of motion-related words relative to the total number of words, while $\log(w + 1)$ quantifies the level of detail using a logarithmic scale.
Figure~\ref{fig3:wordStatistic} (left) illustrates the balanced distribution of words related to motion (red color) and detail (green color). Figure~\ref{fig3:wordStatistic} (right) depicts the length distribution of captions in HMD-270K, which follows a normal distribution with a mean length of approximately 200 words. Additional statistical details are provided in the Appendix.

\section{Methodology}
To further alleviate the issues of omissions and incorrect outputs in captioning motions and details by MLLMs using the HMD-270K dataset, inspired by reinforcement learning (RL), we designed a Caption Set Equivalence Reward (CSER) based on the GRPO algorithm to improve caption correctness and completeness. 
\subsection{Group Relative Policy Optimization}
GRPO~\cite{shao2024deepseekmath} is a variant of Proximal Policy Optimization (PPO)~\cite{schulman2017proximal}, a policy gradient-based RL  
algorithm. 
Unlike PPO, GRPO estimates advantages by comparing candidate answers, without relying on a critic model.
% Unlike PPO, GRPO eliminates the reliance on a critic model and instead estimates advantages by comparing candidate answers. 
Specifically, for a given query $q_i$, GRPO generates multiple distinct candidate outputs $\{o_1, o_2, o_3, \dots, o_n\}$ based on the current policy $\pi_{\theta_{\text{old}}}$. These outputs are then evaluated using a predefined reward function to obtain corresponding rewards $r_1, r_2, \dots, r_n$. The relative quality $A_i$ of these candidate answers is determined by calculating the mean and standard deviation of these rewards:
\begin{equation}
A_i = \frac{r_i - \text{mean}\left(\{r_1, \ldots, r_n\}\right)}{\text{std}\left(\{r_1, \ldots, r_n\}\right)}.
\label{eq.answer}
\end{equation}
GRPO encourages the model to prioritize the responses with higher advantages within the group by updating the policy $\pi_\theta$ using the following clipped surrogate objective:
\begingroup
\small % 缩小字体
\begin{equation}
\begin{split}
\MoveEqLeft[2] \mathcal{J}_{\text{GRPO}}(\theta) = \mathbb{E}_{q,\{o_i\}} \bigg[ \frac{1}{G} \sum_{i=1}^{G} \bigg( 
\min\left( \frac{\pi_\theta(o_i|q)}{\pi_{\theta_{\text{old}}}(o_i|q)} A_i, \right. \\
&\quad \left. \text{clip}\left( \frac{\pi_\theta(o_i|q)}{\pi_{\theta_{\text{old}}}(o_i|q)}, 1-\epsilon, 1+\epsilon \right) A_i \right) \\
&\quad - \beta \; \mathbb{D}_{\text{KL}}\left( \pi_\theta \| \pi_{\text{ref}} \right) \bigg) \bigg],
\end{split}
\label{eq.grpo}
\end{equation}
\endgroup
where $ \mathbb{E}$ means the expectation operation, $G$ Denotes the number of group, \(\text{clip}(\cdot, 1-\epsilon, 1+\epsilon)\) is the clipping function that restricts the input value to the interval \([1-\epsilon, 1+\epsilon]\). $\mathbb{D}_{\text{KL}}$ Stands for Kullback–Leibler Divergence between the new and the reference (old) policies to stabilize updates.
% between the new policy and the old policy to constrain the magnitude of policy updates and ensure stable training.
\begin{figure}[t]
  \centering
  \includegraphics[width=1\linewidth, keepaspectratio]{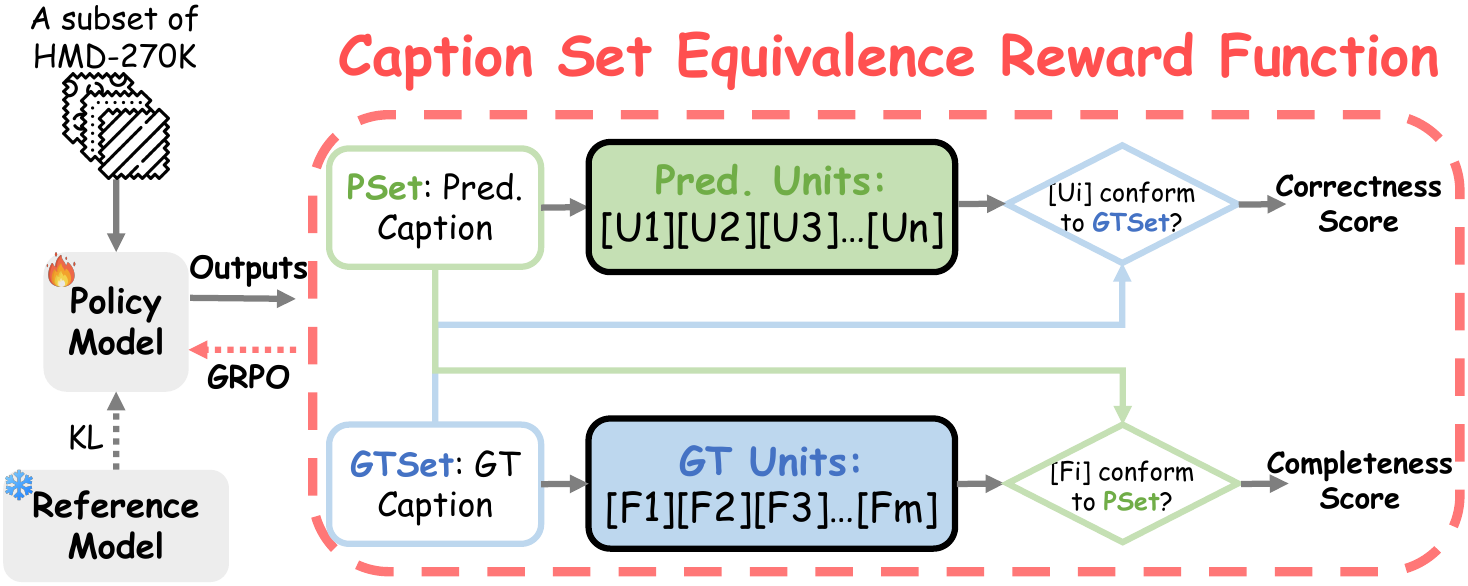}
  % \caption{Illustration of Our GRPO}
  \caption{Caption reinforcement fine-tuning with GRPO.}
  \label{fig3:reward}
\end{figure}

\subsection{Caption Set Equivalence Reward}
The reward function is crucial in reinforcement learning, guiding the model's optimization direction (as shown in Equations~\eqref{eq.answer} and~\eqref{eq.grpo}). To address overlooked video elements and erroneous descriptions in video captioning, we developed the Caption Set Equivalence Reward (CSER) functions. CSER is an elegant approach that ensures the correctness and completeness of the predicted captions by leveraging the set equivalence theory~\cite{cantor1874ueber}, which means that for every element in one set, there is exactly one corresponding element in the other set, and vice versa. Specifically, we introduce two components: Correctness Score ensures that units of the predicted caption set match the GT caption set, thereby preventing erroneous outputs. Meanwhile, Completeness Score ensures that units of the GT caption set are fully represented in the predicted caption set, guaranteeing full coverage of GT information. Together, these scores encourage semantic equivalence between the predicted and GT captions through a unit-to-set matching and bidirectional validation strategy, thereby ensuring the model accurately captures all video elements and matches the captions precisely to the video content.

As shown in Figure \ref{fig3:reward}, aligned with the Fine-Grained Examination stage in the HMD-270K construction pipeline, we decompose both the predicted and GT captions into minimal semantic units using the Qwen3-32B large language model. This decomposition follows the principle of breaking down descriptions into irreducible semantic elements. After decomposition, the predicted caption is represented as a sequence of prediction units: $U_1, U_2,..., U_n$, while the GT caption is transformed into a sequence of GT fact units: $F_1, F_2,..., F_m$. We then assess the generation quality using two scores based on the concept of set equivalence:

\begin{itemize}
    \item \textbf{Correctness Score}: Measured by the matching accuracy between predicted units and the entire GT Caption. If a single predicted unit fails to match the GT Caption as a whole, it indicates that the unit contains information irrelevant to the video content, thereby reducing the overall accuracy of the generated caption.
    \item \textbf{Completeness Score}: Calculated as the proportion of GT facts covered by the entire predicted caption. If a fact in the GT is not covered by the predicted caption as a whole, it signifies that this key part of the video content is not described, thus reducing the overall completeness of the generated caption.
\end{itemize}
The specific formulas are as follows:
\begin{equation}
S_{correctness} = \frac{\sum_{i=1}^{n} \mathbb{I}(U_i \in C_{gt})}{n},
\end{equation}
\begin{equation}
S_{completeness} = \frac{\sum_{j=1}^{m} \mathbb{I}(F_j \in C_{pred})}{m},
\end{equation}
where $C_{pred}$ and $ C_{gt}$ denote the predicted caption and the ground truth caption, respectively; $S_{correctness}$ and $S_{completeness}$ denote the Correctness Score and Completeness  Score; $n$ and $m$ represent the number of units and facts, respectively. 
We use the Qwen3-32B model to assess the relevance of units to the GT caption and the coverage of facts by the predicted caption. 
Notably, we encourage semantic equivalence rather than strict equality~\cite{halmos1960naive} between the predicted and GT caption sets, which aligns with RL optimization goals. Details on unit decomposition and relevance judgment prompts are in the Appendix.

\subsection{Training Processes}
We adopted Qwen2.5-VL-7B as the base model, with the training process divided into two phases: Supervised Fine-Tuning (SFT) and reinforcement learning training. In the first phase, we performed SFT on the HMD-270K dataset, enabling the model to balance motion and detail better. 
In the second phase, we first screened 12K samples from the HMD-270K dataset, all of which achieved 100\% unit accuracy. Each of these samples was then inferred 8 times (Group number) using the stage-one–trained Qwen2.5-VL-7B. The resulting CSER scores were used to compute the within-group variance, and the top 50\% highest-variance samples (6K) were retained as the GRPO training set.
% In the second phase, we first screened 12K samples from HMD-270K whose unit accuracy reached 100\%. Each of these samples was then inferred 8 times (Group number) with the stage-one–trained Qwen2.5-VL-7B, and the resulting CSER scores were used to compute the within-group variance; the top 50\% highest-variance samples (6K) were finally retained as the GRPO training set.
Detailed data filtering methods are provided in the Appendix. This stage of training aims to optimize the model's correctness and completeness, facilitating the generation of higher-quality video captions.
The final reward function for GRPO-based training combines multiple scoring components:
\begin{equation}
\text{Reward} =  S_{\text{format}} +  S_{\text{correctness}} + S_{\text{completeness}}.
\end{equation}
% 数据集如果有bias，0.5可能效果更好？
The format score $S_{\text{format}}$ is to enable the model to output responses in the format we desire. 
% \begin{equation}
% \text{Reward} = \lambda_{\text{f}} S_{\text{format}} + \lambda_{\text{cor}} S_{\text{correctness}} + \lambda_{\text{com}} S_{\text{completeness}}.
% \end{equation}
% % 数据集如果有bias，0.5可能效果更好？
% The format score $S_{\text{format}}$ is to enable the model to output responses in the format we desire. Here, $\lambda_{\text{f}}$, $\lambda_{\text{cor}}$, and $\lambda_{\text{com}}$ represent the corresponding weights for each component.
% \begin{equation}
% Reward = S_{format} + S_{Correctness}  +S_{Completeness}.    
% \end{equation}
% The format score $S_{format}$ is to enable the model to output responses in the format we desire.

\begin{table*}[!ht]
    \centering
    % \small
    % \scriptszie
    \renewcommand{\arraystretch}{0.95}
    \renewcommand{\tabcolsep}{0.44mm}
    % {\fontsize{9}{10}\selectfont
    \begin{tabular}{l|ccccccc}
    \toprule
    \multirow{-2}{*}{\textbf{Model}}&
    % \shortstack{Model\\Metric} &
    % \shortstack{Model\\Metric} &
    \shortstack{Average\\(Acc / Score)} & 
    \shortstack{Detailed\\(Acc / Score)} &
    \shortstack{Camera\\(Acc / Score)} & 
    \shortstack{Short\\(Acc / Score)}&
    \shortstack{Background\\(Acc / Score)}& \shortstack{Object\\(Acc / Score)} & \\
    %     \shortstack{\textbf{Model}\\Metric} & \textbf{\shortstack{Average\\(Acc / Score)}} & 
    % \textbf{\shortstack{Detailed\\(Acc / Score)}} &
    % \textbf{\shortstack{Camera\\(Acc / Score)}} & 
    % \textbf{\shortstack{short\\(Acc / Score)}}&
    % \textbf{\shortstack{Background\\(Acc / Score)}} & \textbf{\shortstack{Object\\(Acc / Score)}} & \\
    \hline
    % \textcolor{lightgreen}
    {\small\textit{\textbf{Video Caption MLLMs}}}\\
ShareGPT4Video-8B~\cite{chen2024sharegpt4video} & 36.2/1.9 & 35.6/1.8 & 33.3/1.8 & 39.1/1.9 & 35.8/1.8 & 37.1/1.9 \\
    Vriptor~\cite{yang2024vript} & 37.7/2.0 & 38.5/2.0 & 37.6/2.0 & 38.4/2.0 & 37.1/1.9 & 37.0/1.9 \\
    AuroraCap-7B~\cite{chai2025auroracap} & 38.2/2.0 & 41.3/2.1 & 43.5/2.3 & 32.1/1.7 & 35.9/1.8 & 39.0/2.0 \\
    % Cockatiel-13B~\cite{qin2025cockatiel} & 43.8/2.3 & 44.4/2.3 & 42.6/2.2 & 43.5/2.3 & 44.1/2.3 & 44.4/2.3 \\
    Tarsier-7B~\cite{wang2024tarsier}&40.2/2.1&38.3/2.1&42.6/2.3&41.7/2.2&36.4/1.9&42.2/2.2&\\
    VideoCap-R1-7B~\cite{meng2025videocap}&43.0/2.3&43.8/2.4&41.7/2.3&35.2/1.9&\textbf{47.2/2.5}&\underline{{47.0/2.5}}\\
    
    \hline
    % \textcolor{lightgreen}
    {\small\textit{\textbf{General MLLM}}}\\
    % VideoChatGPT~\cite{maaz2023video} & 34.2/1.8 & 34.7/1.8 & 33.2/1.7 & 35.7/1.9 & 33.8/1.8 & 33.5/1.8 \\
    VILA-v1.5-8B~\cite{lin2024vila} & 40.5/2.1 & 42.8/2.2 & 39.7/2.1 & 39.3/2.0 & 39.8/2.1 & 40.9/2.1 \\
    VideoChat2-7B~\cite{li2024mvbench} & 36.5/1.9 & 40.5/2.1 & 31.9/1.7 & 40.2/2.1 & 34.9/1.8 & 34.9/1.8 \\
    InternVL-v2-8B~\cite{cai2024internlm2} & 33.7/2.0 & 34.9/1.8 & 39.1/2.1 & 33.0/1.7 & 37.5/1.9 & 44.2/2.2 \\
    % mPLUG-Owl-Video~\cite{ye2024mplug} & 38.9/2.0 & 40.3/2.1 & 38.2/2.0 & 40.2/2.1 & 37.1/1.9 & 38.5/2.0 \\
    % Aria-3.5Bx8~\cite{li2024aria} & 42.5/2.2 & 47.3/2.4 & 42.4/2.2 & 33.2/1.8 & 40.9/2.1 & 48.9/2.5 \\
    LLaVA-OneVision-7B~\cite{li2024llava} & 38.8/2.0 & 41.8/2.2 & 37.6/2.0 & 41.6/2.1 & 34.3/1.8 & 38.8/2.0 \\
    LLaVa-Video-7B~\cite{zhang2024videoinstructiontuningsynthetic}&39.0/2.0&35.0/1.8&\underline{46.1/2.3}&32.8/1.7&37.6/1.9&46.2/2.4\\
    Video-R1-7B~\cite{feng2025video}& \underline{43.9/2.3} & \underline{45.6/2.4} & 42.7/2.2 & \underline{44.5/2.3} & 40.6/2.1 & 45.9/2.3 \\
    \hline
    \cellcolor{ourbaseline}Qwen2.5-VL-7B~\cite{bai2025qwen2} &
    \cellcolor{ourbaseline}42.7/2.2 &
    \cellcolor{ourbaseline}43.4/2.3 &
    \cellcolor{ourbaseline}41.3/2.2&
    \cellcolor{ourbaseline}42.2/2.3 &
    \cellcolor{ourbaseline}41.4/2.1 &
    \cellcolor{ourbaseline}45.2/2.3 \\
    \cellcolor{ourgray}\textbf{OwlCap-7B}&
    \cellcolor{ourgray}\textbf{{46.9/2.4}} &
    \cellcolor{ourgray}\textbf{{46.8/2.4}} &
    \cellcolor{ourgray}\textbf{{46.9/2.5}} &
    \cellcolor{ourgray}\textbf{{47.0/2.4}} & 
    \cellcolor{ourgray}\underline{{46.5/2.4}}&
    \cellcolor{ourgray}\textbf{{47.1/2.5}} \\
    \bottomrule
    \end{tabular}
    % }
    % The best and second-best results are highlighted in \textbf{bold} and \underline{underlined}
    \caption{Quantitative comparison with 12 cutting-edge competitors on the VDC benchmark. Best results in \textbf{bold}, second-best in \underline{underlined}. Qwen2.5-VL-7B serves as the baseline. Higher values indicate better performance for all metrics.}
    \label{tab:vdc}
\end{table*}
\section{Experiments}
% The experimental section of this study consists of four parts:
% \begin{itemize}
%     \item \textbf{Experiment Setups}: Elaborate on the basic configurations of the experiments, including datasets, model parameters, and evaluation criteria.
%     \item \textbf{Main Results}: Compare the performance of different models on three benchmark datasets (VDC, DREAM-1k and VidCapBench-AE) for the cideo captioning task.
%     \item \textbf{Ablation Studies}: Verify the effectiveness of each component (i.e., the proposed dataset construction method and reward function design) through controlled variable experiments.
%     \item \textbf{Downstream Task Validation}: Use captions generated by state-of-the-art video captioning models as prompts for the text-to-video task, and further validate the quality of generated captions by quantifying the differences between the generated videos and the original videos.
% \end{itemize}
\subsection{Experiment Setups}
\textbf{Implementation details.} During video training, the sampling frame rate is fixed at 2 FPS. In the SFT phase, we adopt the Adam optimizer with a learning rate of 1e-5, using a global batch size of 64 for one epoch of training, which takes approximately 26 hours. For the GRPO phase, following Video-R1~\cite{feng2025video}, we set $\beta$ = 0.001 for the KL penalty. Each machine utilizes 6 GPUs for training and 2 dedicated GPUs for reward inference, with a global batch size of 24. Training for one epoch in this phase takes about 35 hours. 
All experiments are implemented on 32 H20 (96GB) GPUs, with additional experimental details provided in the Appendix.\\
\textbf{Benchmarks.}
We evaluate our model using two mainstream benchmark datasets: Video Detailed Captions (VDC)~\cite{chai2025auroracap} and DREAM-1K~\cite{wang2024tarsier}.
The VDC benchmark, comprising over 1,000 videos, rigorously evaluates detailed video caption quality, focusing on characterizing video details. DREAM-1K assesses models' ability to capture fine-grained actions and events, emphasizing accurate motion descriptions.
For fair comparison, all benchmark tests and metrics are strictly implemented according to the original code from the papers, and comparative model data is sourced from official leaderboards. We further conduct experiments on the T2V-oriented VidCapBench-AE~\cite{chen2025VidCapBench} and the fine-grained CaReBench~\cite{xu2024carebench} to demonstrate generalizability.
% For fair comparison, all benchmark tests and metrics are conducted strictly following the code implementation in the original papers, and comparative model data is sourced from official leaderboards. 
% We additionally carry out experiments on the T2V-oriented VidCapBench-AE~\cite{chen2025VidCapBench} and the fine-grained CaReBench~\cite{xu2024carebench} to demonstrate generalizability.
% We also conduct experiments on multidimensional VidCapBench-AE~\cite{chen2025VidCapBench} and CaReBench~\cite{xu2024carebench} benchmarks to demonstrate generalizability. 
Details on the benchmarks and evaluation metrics are in the Appendix.\\
% More details on benchmarks and metrics are provided in the Appendix.\\
% , and VidCapBench-AE~\cite{chen2025VidCapBench}. 
% The VDC benchmark, consisting of over 1,000 videos, enables rigorous assessment of detailed video caption quality, with a focus on characterizing video details. DREAM-1K is specifically designed to evaluate the model's capability to capture fine-grained actions and events in videos, emphasizing the accuracy of motion descriptions. 
% VidCapBench-AE is a video caption evaluation scheme specifically designed for T2V task, agnostic to any particular caption format.
% More benchmark and metrics details in Appendix.\\
% Datasets and metrics details are provided in the Appendix.\\
% 指标还是
% \textbf{Evaluation Metrics.}
% Benchmark-specific metrics are employed to assess the performance of video caption models. For VDC, we utilize the Accuracy (Acc~$\uparrow$) and VDCscore (Score~$\uparrow$) metrics to evaluate performance across five dimensions (Detailed, Camera, Short, Background, and Object) and their average value, Average; for DREAM-1K, the F1~$\uparrow$, Pecision (P~$\uparrow$), and Recall(R~$\uparrow$) metrics are included; correspondingly, the  Accuracy (Acc~$\uparrow$), Precision (Pre~$\uparrow$) and Coverage (Cov~$\uparrow$) metrics are used to evaluate VidCapBench-AE. Additionally, to compare the performance of caption models for T2V tasks, we employ the popular metrics Structural Similarity Index (SSMI~$\uparrow$), Peak Signal to Noise Ratio (PSNR~$\uparrow$), and Fréchet Inception Distance (FID~$\downarrow$) to assess the quality of generated videos.
\begin{table}[!t]  % 
    \centering
    % \scriptsize
    \renewcommand{\arraystretch}{0.95}
    \renewcommand{\tabcolsep}{2.6mm}
    \begin{tabular}{l|cc}  % 单列结构（模型名 + 平均性能）
    \toprule
     \multirow{2}{*}{\textbf{Model}}&  \multicolumn{2}{c}{{Overall}} \\
     & F1 & Recall \\
    \hline
    % \textcolor{lightgreen}
    {\small\textit{\textbf{Video Caption MLLMs}}}\\
    ShareGPT4Video~\cite{chen2024sharegpt4video} & 19.5  & 15.4\\
    % Vript&&\\
    AuroraCap-7B~\cite{chai2025auroracap}&20.8  & 18.1\\
    Tarsier-7B~\cite{wang2024tarsier}&\underline{34.6}  & 30.2\\
    VideoCap-R1-7B~\cite{meng2025videocap}& 34.2  & \underline{34.7}\\
    \hline
    % \textcolor{lightgreen}
    {\small\textit{\textbf{General MLLM}}}\\
    % Video-LLaVa-7B~\cite{lin2023video} & 20.4 / 28.1 / 16.0 \\
    VILA-v1.5-8B~\cite{lin2023vila}&29.9  & 25.8\\
    VideoChat2-7B~\cite{li2023videochat}&26.6 & 23.3 \\
    % VideoLLaMA2-7B~\cite{cheng2024videollama} & 26.2 / 31.5 / 22.4 \\
    % LLaVA-OV-7B~\cite{li2024llava} & 31.7 / 34.3 / 29.4 \\
    InternVL2-8B~\cite{cai2024internlm2} & 26.9  & 24.7 \\
    LLaVa-OneVision~\cite{li2024llava}&31.7  & 29.4\\
    LLaVA-Video-7B~\cite{zhang2024videoinstructiontuningsynthetic} & 32.5 & 28.4\\
    Video-R1-7B~\cite{feng2025video}&33.3  & 34.5\\
    % VideoChar-R1&34.4&33.6&35.2\\
    \hline
    \cellcolor{ourbaseline}Qwen2.5-VL-7B~\cite{bai2025qwen2}&
    \cellcolor{ourbaseline}30.1  &
    \cellcolor{ourbaseline}29.7 \\
    \cellcolor{ourgray}\textbf{OwlCap} & 
    \cellcolor{ourgray}\textbf{34.7} & \cellcolor{ourgray}\textbf{35.3} \\
    \bottomrule
    \end{tabular}
    \caption{Evaluation results on DREAM-1K benchmark.}
    \label{tab:DREAM-1k}
\end{table}
\subsection{Main Results}
% \subsection{Evaluation of Owlcap}
% 后面加点数据 xxxacheive 46.3
\textbf{Result on video captioning.}
We conducted a comprehensive evaluation of OwlCap across two distinct benchmarks, comparing its performance with that of other general MLLMs and specialized video captioning MLLMs. As clearly illustrated in Table~\ref{tab:vdc}, OwlCap, after undergoing the two-stage training process leveraging the HMD-270K dataset, has achieved a visible improvement in every aspect of the VDC benchmark when compared to the baseline model. Meanwhile, it also outperforms the existing general-purpose MLLMs and video captioning MLLMs, demonstrating OwlCap’s advantage in detailed description. 
Similarly, as presented in Table~\ref{tab:DREAM-1k}, within the DREAM-1K benchmark, which focuses on motion description, OwlCap has also achieved a significant improvement compared to the Baseline. This consistent performance enhancement further serves to validate the effectiveness of OwlCap. More detailed results, as well as other benchmark experimental results, are provided in the Appendix.
% the HMD-270K dataset, and CSER
% as well as the utility of the Completeness and Correctness Scores 
% in guiding model optimization. 
% Finally, as shown in Table~\ref{tab:VidCapBench-AE results}, in VidCapBench-AE, a benchmark specifically designed to assess the quality of T2V captions, OwlCap has also shown a notable improvement, rounding out a strong overall performance across all evaluated metrics.
\begin{table}[!t]  % 
    \centering
    \renewcommand{\arraystretch}{0.95}
    \renewcommand{\tabcolsep}{3.5mm}
    \begin{tabular}{l|ccc}  % 调整列结构，增加三列
    \toprule
    Model  
    & SSMI~$\uparrow$ & PSNR~$\uparrow$&FID$~\downarrow$  \\
    \hline
    Tarsier-7B &0.30
     &28.45
      &242.35\\
    AuroraCap-7B &
     0.27&25.97
    &261.45  \\
    VideoCap-R1-7B&0.30&28.60&240.72\\
    \hline
    \cellcolor{ourbaseline}Qwen2.5-VL-7B &
    \cellcolor{ourbaseline}0.29&
   \cellcolor{ourbaseline}28.02&
   \cellcolor{ourbaseline}245.09 \\
    \cellcolor{ourgray}\textbf{OwlCap} &
    \cellcolor{ourgray}\textbf{0.33}&
    \cellcolor{ourgray}\textbf{29.32}&
     \cellcolor{ourgray}\textbf{231.60}\\
    \bottomrule
    \end{tabular}
    \caption{Comparison of model caption effects on T2V task.}
    % \caption{Comparative results of contrastive models in T2V task.}
    \label{tab:T2V}
\end{table} % 用于更美观的横线

% \textbf{Validating video captioning models by T2V quality evaluation.}
\textbf{Validating video captioning models via T2V evaluation.}
To further validate OwlCap's performance, we conducted additional evaluations using the T2V task, a downstream application of video captioning.
Specifically, we collected 200 videos (5–30 seconds long) from Pixabay and used Tarsier, AuroraCap, VideoCap-R1, Qwen2.5-VL-7B, and OwlCap to generate captions. These captions were then fed into HunyuanVideo~\cite{kong2024hunyuanvideo} to generate corresponding videos.
Subsequently, we compared each generated video with its corresponding reference by uniformly sampling 30 frames from both and then computing the evaluation metrics—SSIM, PSNR, and FID—on these aligned frames.
As shown in Table~\ref{tab:T2V}, OwlCap demonstrated significant advantages in the T2V task: videos generated from its captions exhibited notably higher quality than those from competing models. 
See Appendix for additional visual comparisons.

\subsection{Ablation Studies}
% \textbf{Effect of MDF and MEG in HMD-270K construction pipeline.} 
\textbf{Effect of MDF and FGE.} 
To validate the effectiveness of the Motion-Detail Fusion (MDF) and Fine-Grained Examination (FGE) stages in the HMD-270K construction pipeline, we selected 4K videos from the HMD-270K dataset and generated corresponding captions using Tarsier (the motion-oriented model) and Qwen2.5-VL-72B (the detail-completion model) for comparative analysis. 
Inspired by~\cite{yan2024vigor}, we evaluated the Motion and Detail Scores of captions using GPT-4o with specific instructions (see Appendix).
% Inspired by~\cite{yan2024vigor}, we evaluated the Motion Score and Detail Score of the captions by providing GPT-4o with the corresponding instructions. The specific prompt can be found in the Appendix.
As shown in Table~\ref{tab:HMD}, the captions generated after Motion-Detail Fusion achieved higher scores in both motion and detail dimensions. 
For Unit Acc, we use GPT-4o in place of InternVL2.5-78B as the judge model for evaluation.
The initial performance in Unit Acc was unsatisfactory. After optimization through Fine-Grained Examination, Unit Acc significantly improved, along with enhancements in both Motion Score and Detail Score.
\\
% 需要重新阐述
\textbf{Effect of HMD-270K and caption score in CSER.}
Ablation experiments in Table~\ref{tab:abla} demonstrate how HMD-270K and the correctness and completeness scores in CSER boost video captioning. Key findings are: (i) First-stage SFT training alone offers limited gains; (ii) RL training without prior SFT on HMD-270K also yields minimal improvement; whereas (iii) using the SFT-trained model as the base for RL achieves significant performance enhancements.
\begin{table}[t]
  \centering
  % \small
    \renewcommand{\arraystretch}{0.95}
    \renewcommand{\tabcolsep}{1.1mm}
  \begin{tabular}{l|cc|c}
    \toprule
        &Motion Score&{Detail Score} & {Unit Acc}  \\
       
    % &Motion&{Detail} & {Unit}  \\
    %    & {Score}&{Score} & {Acc}\\
      % & \textbf{Motion}&\textbf{Detail} & \textbf{Unit}  \\
      %  & \textbf{Score}&\textbf{Score} & \textbf{Acc}\\
    \hline
    Tarsier    &7.32& 6.15 & -  \\
    Qwen2.5-VL-72B   &6.99  &6.36  & -  \\
        MDF   &7.80  &7.02 &78.93   \\
    \cellcolor{ourgray}FGE &
    % Motion-Detail Fusion   &7.80  &7.02 &78.93   \\
    % \cellcolor{ourpink}Fine-Grained Exam. &
    \cellcolor{ourgray}\textbf{7.92}    &
    \cellcolor{ourgray}\textbf{7.13}  &
    \cellcolor{ourgray}\textbf{96.54}  \\
    \bottomrule
  \end{tabular}
    % \caption{Effect of two HMD-270 construction stage.}
  % \caption{Impact of the MDF and FGE stages.}
    \caption{Impact of the MDF and FGE in HMD-270K construction pipeline.}
    % \caption{Effect of the MDF and FGE in HMD-270K construction pipeline.}
  \label{tab:HMD}
\end{table}
\begin{table}[t]
  \centering
  % \small
    \renewcommand{\arraystretch}{0.95}
  \renewcommand{\tabcolsep}{2.2mm}
  \begin{tabular}{c|cc|cc|cc}
    \toprule
          {HMD} & \multicolumn{2}{c|}{{Reward}} &  \multicolumn{2}{c|}{VDC} & \multicolumn{2}{c}{DREAM-1K}\\
     {SFT} & {Com.} & {Cor.} & {Acc}&Score & {F1}  & {Recall}   \\
     
     %  \textbf{HMD} & \multicolumn{2}{c|}{\textbf{Reward}} &  \textbf{VDC}& \textbf{DREAM.} & \textbf{VidCap.} \\
     % \textbf{SFT} & \textbf{Com.} & \textbf{Cor.} & \textbf{Acc.} & \textbf{F1}  & \textbf{Acc.}   \\
    \hline
       &  &  & 42.7 & 2.21&30.1 & 29.7  \\
    \hline
     \checkmark &  &  & 44.8 &2.31& 31.9 &32.1  \\
    \hline
       &\checkmark  &  &43.5 &2.27 &32.0  &31.7    \\
   &  &\checkmark  &43.7 &2.26 &31.7  &31.3    \\
    &\checkmark  &\checkmark  &44.2 &2.29 &32.7  &32.9    \\
    \hline
      \checkmark & \checkmark &  &45.9&2.34  &33.5 &33.9     \\
 \checkmark &  & \checkmark &46.2 &2.35 &33.7 &34.0 \\
     \cellcolor{ourgray}\checkmark &
    \cellcolor{ourgray} \checkmark & 
     \cellcolor{ourgray}\checkmark& 
     \cellcolor{ourgray}\textbf{46.9} & 
     \cellcolor{ourgray}\textbf{2.43} & 
     \cellcolor{ourgray}\textbf{34.7}&
     \cellcolor{ourgray}\textbf{35.3}
     \\
    \bottomrule
  \end{tabular}
  \caption{Contribution of HMD-270K (HMD) SFT, Completeness (Com.), and Correctness (Cor.) score rewards.}
  \label{tab:abla}
\end{table}
\\
\textbf{Comparison of other Caption reward.}
To compare CSER with the Event Score in VideoCap-R1~\cite{meng2025videocap}.
We conducted comparative experiments under the same base model. As shown in Table~\ref{tab:different_reward}, we used the model fine-tuned on HMD-270K via SFT as the baseline, tested the performance of the model on the VDC and DREAM-1K after adding Event Score training. The results indicate that Event Score achieves significant gains on DREAM-1K but only limited improvements on VDC. In contrast, CSER consistently outperforms Event Score across all metrics.
\\
\textbf{Performance of diverse base models.} OwlCap is compatible with multiple Qwen-family models. To investigate the impact of different base models on OwlCap, we present the performance of OwlCap equipped with Qwen2-VL-7B, Qwen2.5-VL-3B, and Qwen2.5-VL-7B in Table~\ref{tab:different_baseline}. It can be clearly seen that, OwlCap consistently outperforms the baselines on both the VDC and DREAM-1K benchmarks, highlighting its stability and effectiveness.
\begin{table}[t]
  \centering
  \renewcommand{\arraystretch}{0.95}
  \begin{tabular}{l|c|cc|cc}
    \toprule
    \multirow{2}{*}{{Method}}
    &Train& \multicolumn{2}{c|}{VDC} & \multicolumn{2}{c}{DREAM-1K} \\
    &Time& Acc &Score&F1&Recall\\
    \hline
   % Baseline & 44.3 & 2.30 &31.5&30.9 \\
    Baseline &-& 44.8 & 2.31 &31.9&32.1 \\
   + Event Score&+33h&45.2  &2.33&32.7&33.2  \\
    \rowcolor{ourgray}
    + CSER&+35h &\textbf{46.9} & \textbf{2.43} & \textbf{34.7} & \textbf{35.3} \\
    \bottomrule
  \end{tabular}  \caption{Comparisons of different caption reward. Event Score is the caption reward proposed in VideoCap-R1.}
    % \caption{Impact of different caption reward.}
\label{tab:different_reward}
\end{table}
\begin{table}[!t]
  \centering
  \renewcommand{\arraystretch}{0.95}
  \renewcommand{\tabcolsep}{1mm}
  \begin{tabular}{l|ll|ll}
  % \begin{tabular}{l|cc|cc}
    \toprule
    \multirow{2}{*}{{Model}}
    & \multicolumn{2}{c|}{VDC} & \multicolumn{2}{c}{DREAM-1K} \\
    & Acc &Score&F1&Recall\\
    \hline
    Qwen2-VL-7B & 39.8 & 2.1 &29.6&26.3 \\
    \rowcolor{ourgray}
    \textbf{OwlCap }& 45.2\textbf{\scriptsize(+5.4)}
    & 2.4\textbf{\scriptsize(+0.3)}
    & 34.4\textbf{\scriptsize(+4.8)} 
    & 34.7\textbf{\scriptsize(+8.4)}
    \\
    \hline
    Qwen2.5-VL-3B & 41.1 & 2.1&29.3&28.9  \\
    \rowcolor{ourgray}
    \textbf{OwlCap }
    & 45.7\textbf{\scriptsize(+4.6)}
    & 2.3\textbf{\scriptsize(+0.2)}
    & 32.5\textbf{\scriptsize(+3.2)}
    & 33.3\textbf{\scriptsize(+4.4)}
    \\
    \hline
    Qwen2.5-VL-7B & 42.7&2.2 & 30.1&29.7  \\
    \rowcolor{ourgray}
    \textbf{OwlCap }
    & 46.9\textbf{\scriptsize(+4.2)}
    & 2.4\textbf{\scriptsize(+0.2)}
    & 34.7\textbf{\scriptsize(+4.6)}
    & 35.3\textbf{\scriptsize(+5.6)}
    \\
    \bottomrule
  \end{tabular}
  \caption{OwlCap performance with Qwen family models.}
  \label{tab:different_baseline}
\end{table}
\begin{figure}[!t]
  \centering
  \includegraphics[width=0.95\linewidth, keepaspectratio]{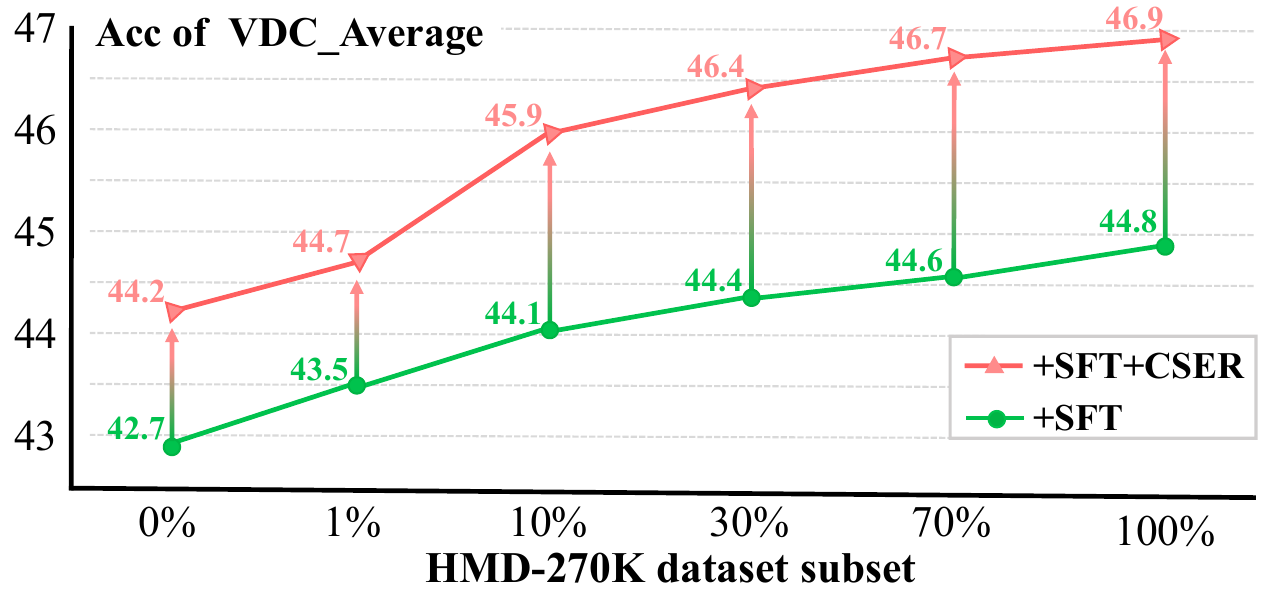}
  \caption{Impact of different ratios in the HMD-270K.}
  \label{fig3:differsize}
\end{figure}
\\
\textbf{Ablation study about different data sizes.} 
Figure~\ref{fig3:differsize} shows the impact of different subsets of HMD-270K with varying proportions on OwlCap. 
We highlight two key findings: first, the performance improvements become increasingly marginal as the data radio grows; second, SFT training serves as the foundation that enables the subsequent GRPO phase based on CESR to yield further improvements.
% It can be seen from the figure that as the sampling ratio gradually increases, the model's performance on the VDC benchmark steadily improves, indicating the high quality and stability of the caption annotations generated by our data construction pipeline.
\section{Conclusion}
The primary purpose of this paper is to explore how to enhance the quality of video descriptions in video captioning. We observe that existing methods struggle to simultaneously capture both dynamic actions and fine-grained details in videos. To address this challenge, we propose a video captioning dataset, Harmonizing Motion-Detail
270K (HMD-270K) via a two-stage pipeline. This framework integrates a  Motion-Detail Fusion (MDF) and a Fine-Grained Examination
(FGE), to synthesize and refine video captions.
Furthermore, to fully leverage the potential of HMD-270K, we introduce a Group Relative Policy Optimization (GRPO)-based reinforcement learning strategy, enhanced with a Caption Set Equivalence Reward (CSER), to improve the correctness and completeness of video captioning capabilities for MLLMs.
By leveraging HMD-270K for SFT pre-training and GRPO post-training with CSER, we develop OwlCap, a model that effectively captures both motion dynamics and detailed descriptions in videos, overcoming the inherent caption-bias limitation in conventional approaches.
\bibliography{aaai2026}
% \clearpage
% \input{supp}
\end{document}